\documentclass[9pt,twocolumn,twoside]{opticajnl}
\journal{opticajournal} % use for journal or Optica Open submissions

\setboolean{shortarticle}{true}
\usepackage{lineno}
\newenvironment{subfigure}{\begin{minipage}{\linewidth}}{\end{minipage}}
\definecolor{todocolor}{RGB}{200, 30, 30}

\title{Foundation-Model Residuals as Signals of Urban Bus Irregularity\thanks{Accepted as a poster at the NetMob 2026 Data Challenge but not presented.}}
\author[1,*]{Csaba I. Sidló}
\author[1]{Jamil J. Ghazal}
\author[1]{Domonkos Rózsay}
\author[1]{Rita Rónai}
\author[1]{András J. Molnár}

\affil[1]{Institute for Computer Science and Control, Hungarian Research Network, Hungary} 
\affil[*]{sidlo@sztaki.hu}
\setboolean{displaycopyright}{false} % Do not include copyright or licensing information in submission.

\begin{document}

\maketitle

Pretrained trajectory models encode mobility patterns learned across cities, but their errors when transferred to a new city are usually treated only as performance limitations.
We instead ask whether these residuals reveal systematic local conditions. 
Architectures such as UniTraj~\cite{NEURIPS2025_3c215225} and TrajFM~\cite{lin2024trajfm} learn transferable trajectory representations, although cross-context transfer remains challenging~\cite{SongGaoIssues}.
Using the NetMob26 bus location, ticketing, weather, network, and socioeconomic data for Niter\'oi~\cite{netmob26}, we analyze spatial and travel-time residuals of these recent trajectory foundation models.
We test whether high residuals concentrate spatially, whether spatial and travel-time residuals overlap, and which urban-context variables explain their distribution.
%The NetMob 2026 Data Challenge \cite{netmob26} provides a rich framework to analyze these dynamics of Niter\'oi, where dense central corridors, peripheral neighborhoods, coastal and hilly terrain, and ferry-oriented access patterns create highly heterogeneous movement contexts.
%offering high-frequency bus GPS traces, ticketing records, route and stop data, meteorological variables, and urban infrastructure layers for Niter\'oi~\cite{netmob26}. 

We interpret high residuals as candidate proxy signals of transit irregularity rather than direct evidence of service disruption, and analyze their contextual associations using Explainable AI (XAI)~\cite{lundberg2017shap}.
Insights into such varied environments are increasingly valuable for evaluating new urban paradigms; for instance, the 15-minute-city perspective ~\cite{moreno15minutecity} serves as a key motivation for our approach, guiding our methods and expanding the analysis beyond standard mobility metrics.
%What looks unusual when a city's transit system is viewed through the lens of a global mobility prior? 

\section{Data and Analytical Methodology}

We first process and extend the NetMob dataset, then generate behavioral benchmarks using pretrained foundation models, and subsequently apply XAI to analyze the systemic drivers behind the prediction residuals.
%\footnote{Derived outputs and analysis code are available from the authors upon request.}

The Niter\'oi dataset contains GPS traces, ticketing, route, weather, and socioeconomic data for spatial cells.
Points-of-interest (POIs) are added with 15-minute-city indicators, counting POI categories within 15-minute walk. Schedule-based metrics are omitted due to widespread data gaps. 

%Trajectory irregularities are detected using outputs from two trajectory foundation models. 
%For comparison, we implement local baselines using linear models, OLS, RIDGE and LASSO with default parameters (and using a: only spatial data, and b: using spatial and time data) targeting the identical trajectory-position and trip-time tasks.
%The basic linear models with RMSE yielded around the same or better results from these.
For comparison, we trained task-matched local baselines exclusively on the Niterói data. For trajectory reconstruction we used linear regression, Ridge and Lasso models based on recent spatial coordinates, with and without temporal features. For travel-time prediction we used the same locally trained regression family. 
Default hyperparameters were used, as these models serve as simple local references.
%Hyperparameters were kept at standard/default settings, as these models serve as simple local reference methods rather than extensively tuned competitors.

\begin{figure}[ht]
\centering
\includegraphics[width=\linewidth]{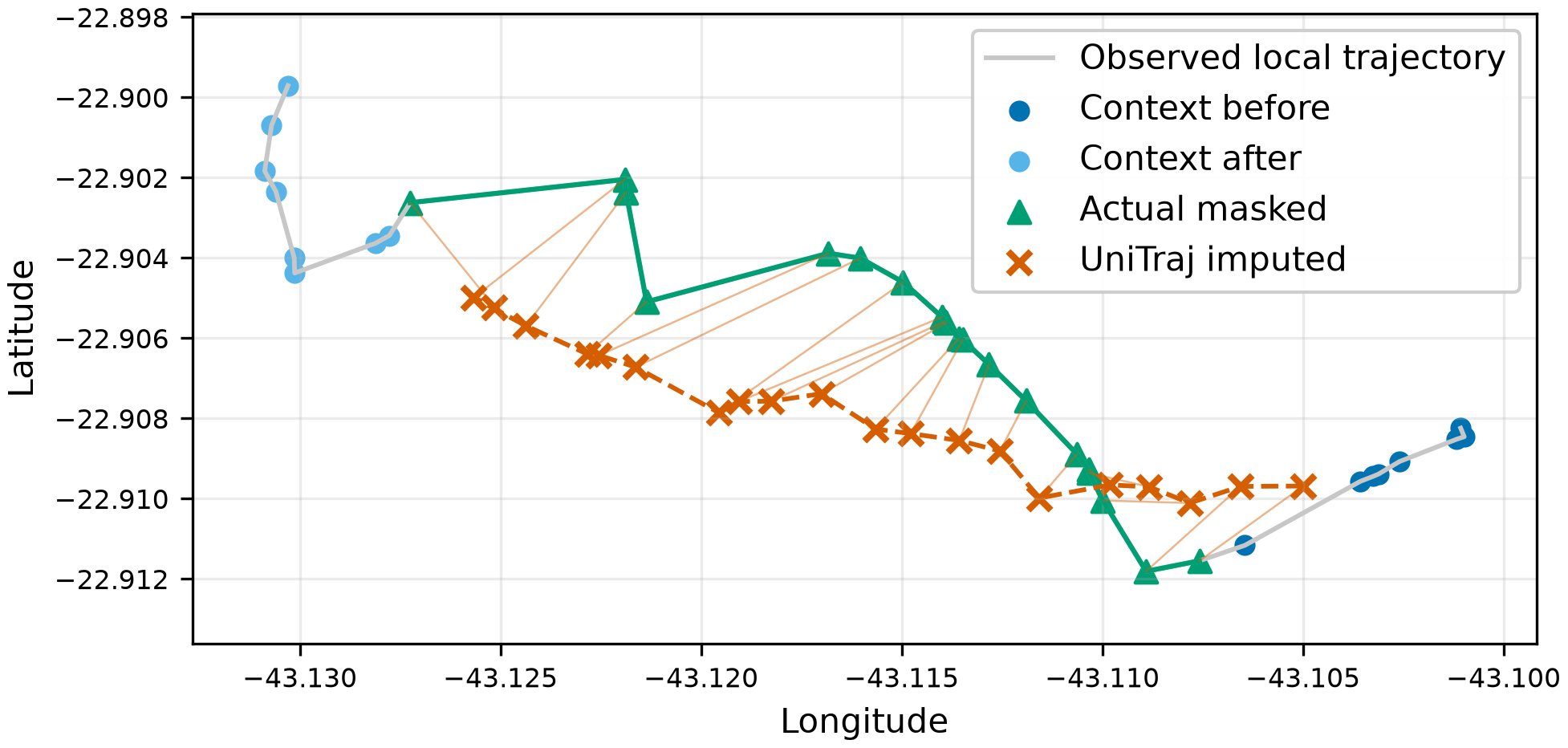}
\caption{Example UniTraj masked-segment imputation.}
\label{fig:unitraj-imputation-example}
\end{figure}

Pretrained TrajFM~\cite{lin2024trajfm} is evaluated for travel-time estimation, with errors measured in minutes.
We evaluate both zero-shot and locally fine-tuned UniTraj~\cite{NEURIPS2025_3c215225} for spatial trajectory reconstruction.
Comparative testing between sliding-window prefix continuation and masked segment reconstruction showed that the masked approach yielded superior performance.
The final setup used windows with 200 locations, 30 points masked or predicted.
Figure~\ref{fig:unitraj-imputation-example} illustrates the setup: UniTraj reconstructs a hidden segment of an observed bus trajectory, and the residual distance becomes the local irregularity signal.

Features and predictions are aggregated to $1,236$ distinct spatial H3 resolution-9 hexagonal cells (with H3 resolution 10 used for sensitivity analysis). 
After filtering out unique identifiers, target variables, raw counts, center-distance proxies, and uninformative generic aggregates, the final explanatory model retains $39$ contextual features with five distinct thematic families: transit demand and rider composition, localized speed and congestion, road network geometry, 15-minute city accessibility indicators, and census-derived socioeconomic status (SES).

%~\cite{chen2016xgboost}
We explain per-cell errors with XGBoost using urban features as inputs and aggregated model error as target.
SHAP values~\cite{lundberg2017shap} provide global importance and, for each cell, the dominant feature family behind the local irregularity score.
We sum absolute SHAP contributions within predefined feature families and assign each mapped cell to the family with the largest total absolute contribution. 
This way we visualize local explanatory context, not only a citywide ranking.
To reduce target circularity, we omit closely related speed-derived variables from one explanatory specification; remaining speed/congestion features are operationally useful but are treated as associative rather than independent causal evidence.

\section{Results and Discussion}
We evaluated both the zero-shot pretrained and locally fine-tuned UniTraj variants and retained the fine-tuned model, which achieved lower reconstruction error, for the subsequent residual and hotspot analyses.
At the GPS-point level, UniTraj spatial imputation errors average $421$\,m. 
%rising to $521$\,m after H3 cell aggregation. 
For comparison, a locally trained task-matched baseline achieved a mean trajectory-position error of $664$\,m. %while its full-trip travel time prediction yields a mean error of $5.94$\,min on average $54$-minute trips. 
%  (median: $566$\,m)
UniTraj residuals are non-uniformly distributed, decreasing with population density (Spearman $\rho = -0.37, p < 0.001$) and clustering along structurally complex transit corridors.
Under random 5-fold cross-validation, the XGBoost meta-model explains  $R^2 \approx 0.55$ of cell-level UniTraj error.
This predictability indicates that UniTraj residuals have a substantial spatial association with the measured urban context.
%This predictability suggests that UniTraj residuals are systematically associated with localized urban context.

Across $7,092$ test trips, TrajFM achieved an MAE of $5.12$ min and a MAPE of $9.47\%$, outperforming the best local baseline MAE of $5.94$ min.
%For travel-time prediction, TrajFM achieved an MAE of $5.12$ min, an RMSE of $7.60$ min, and a median absolute error of $3.56$ min. 
Its mean bias of $-1.14$ min indicates a modest tendency to underestimate trip duration. 

The TrajFM meta-model has limited explanatory power (\(R^2 \approx 0.07\)); its SHAP results should therefore be interpreted cautiously. 
Within this weakly predictive model, passenger-demand features (free-pass ridership share) show the largest associations, followed by 15-minute accessibility and congestion metrics.
%TrajFM's travel-time errors increase more moderately toward the periphery ($\rho \approx +0.27$), and the meta-model yields lower explanatory power ($R^2 \approx 0.07$). 
%TrajFM's SHAP profile is dominated by passenger-demand features, specifically free-pass ridership share, followed by 15-minute accessibility and localized congestion metrics. 

\begin{figure}[t]
  \centering
  \begin{subfigure}
    \includegraphics[width=\linewidth]{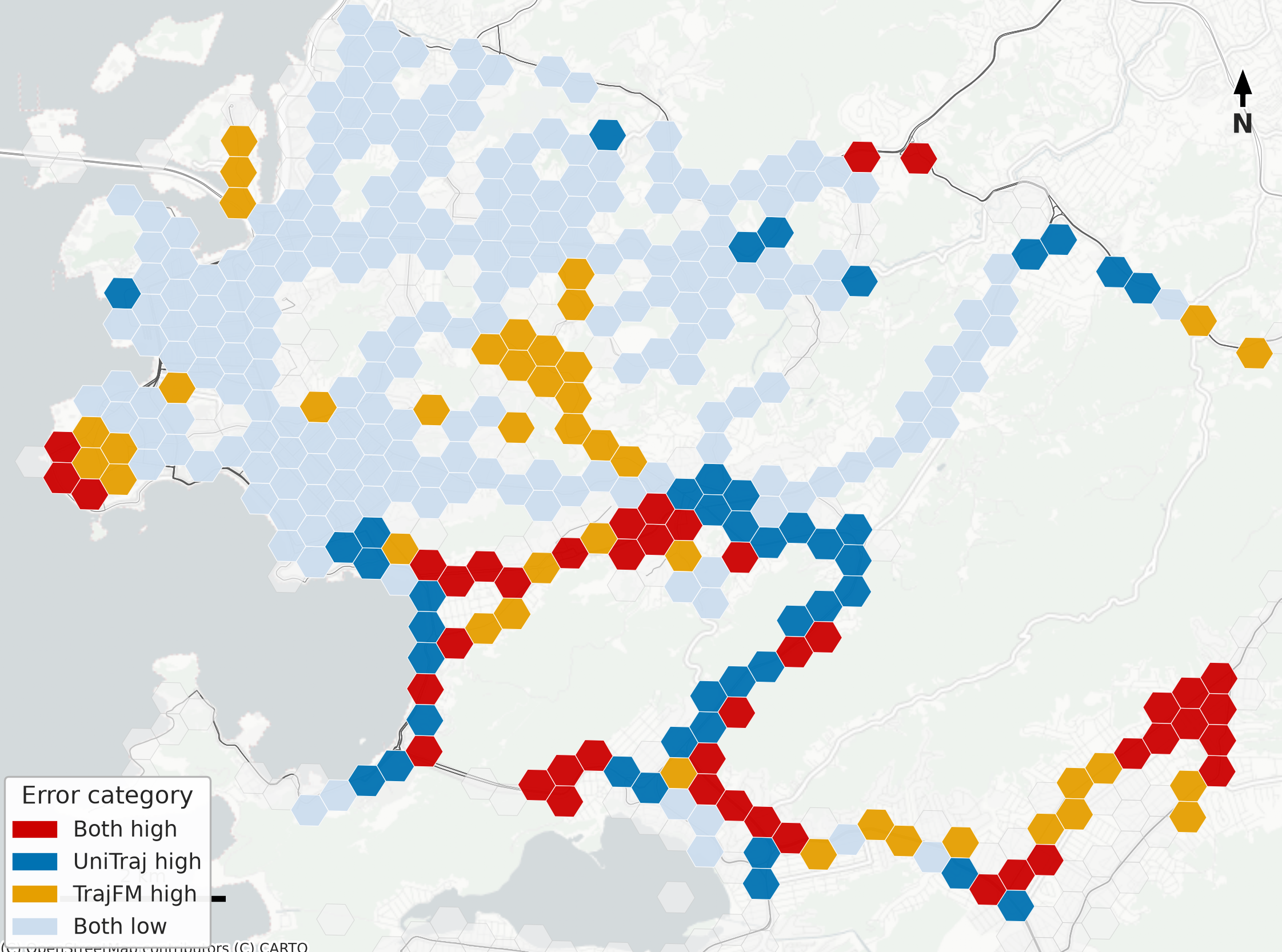}
    \caption{Irregularity hotspots: highlighted cells fall within the top quartile for UniTraj spatial or TrajFM travel-time error.}
    \label{fig:joint-hotspots}
  \end{subfigure}
  \begin{subfigure}
    \includegraphics[width=\linewidth]{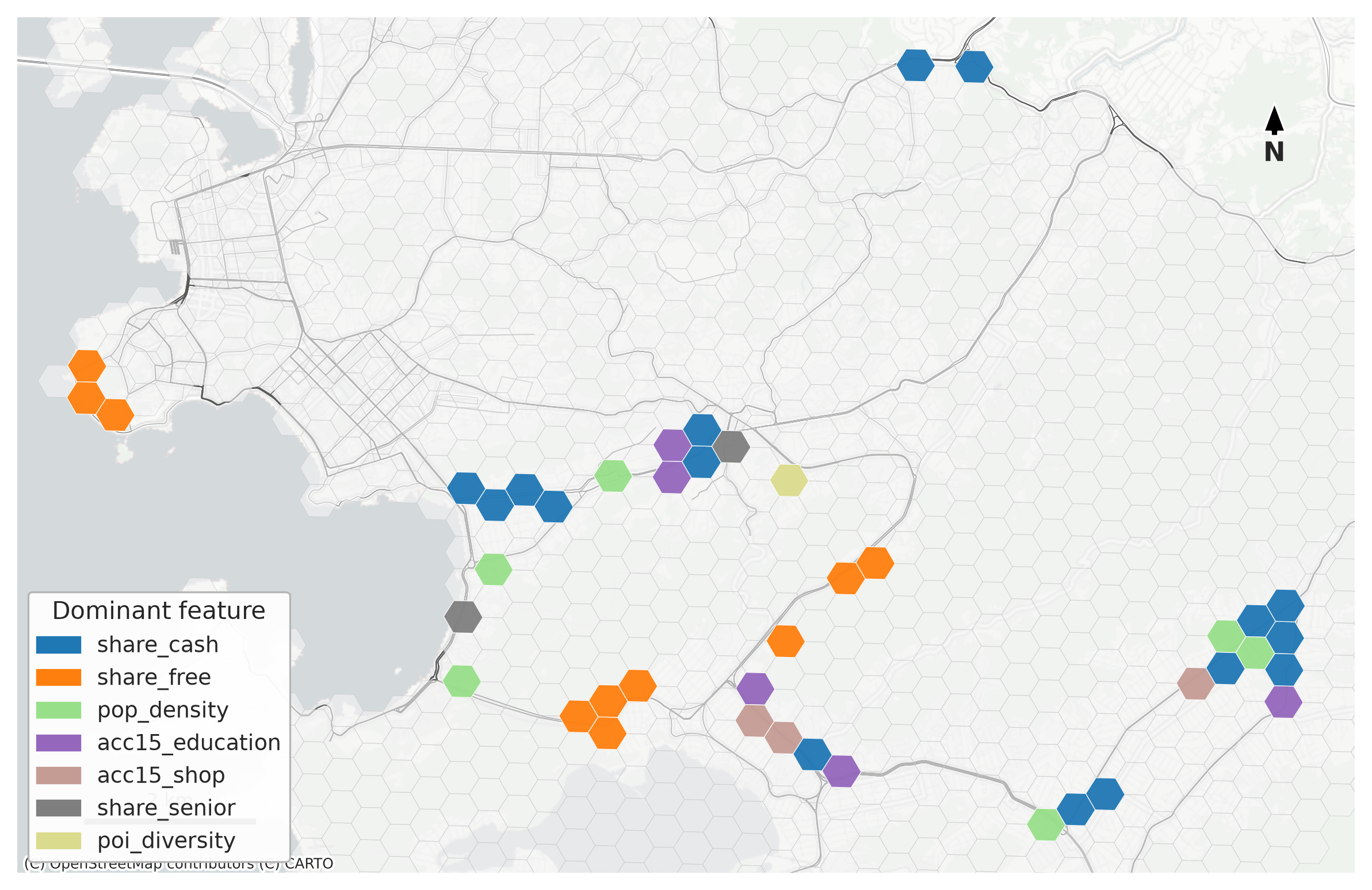}
    \caption{Dominant SHAP feature family within joint-hotspot cells.}
    \label{fig:both-high-driver-map}
  \end{subfigure}
\end{figure}

%UniTraj and TrajFM cell-level errors are moderately correlated ($\rho \approx +0.50$), overlapping in $39$ cells within their respective top quartiles - we call these ``joint hotspots'', as illustrated in Fig.~\ref{fig:joint-hotspots}.
UniTraj and TrajFM cell-level errors are moderately correlated ($\rho \approx +0.50$). 
Their respective top-quartile hotspot sets overlap in 39 cells, referred as ``joint hotspots`` (Fig.~\ref{fig:joint-hotspots}).
Hotspot locations remained stable across alternative feature subsets, H3 cell resolutions, and trajectory-masking strategies, indicating that the spatial patterns were not artifacts of a modeling configuration.

\begin{figure}[t]
  \centering
  \begin{subfigure}
    \includegraphics[width=\linewidth]{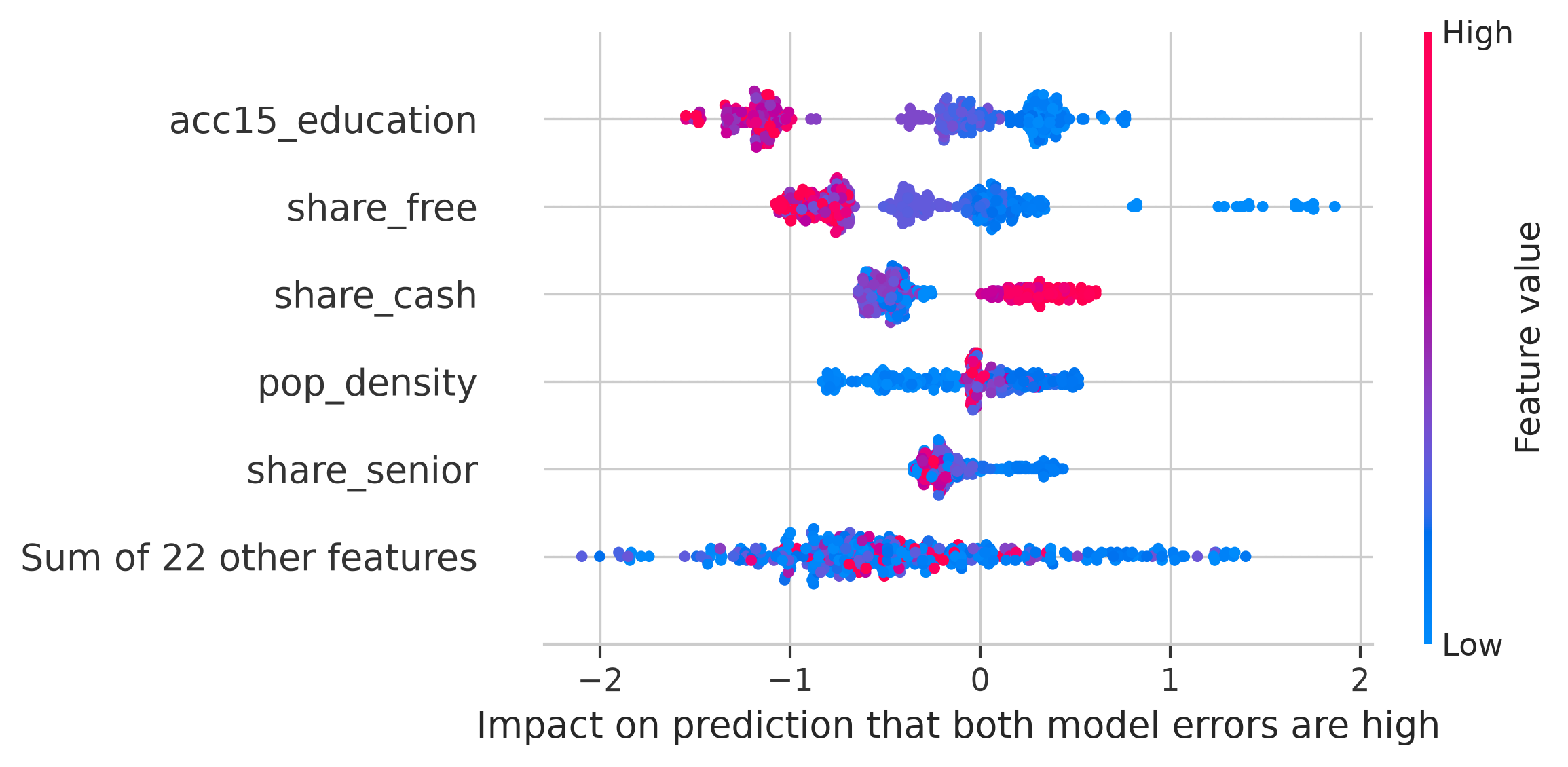}
    \caption{SHAP beeswarm with the top features for the joint hotspot class.     Higher SHAP values indicate a greater predicted probability of joint-hotspot membership.}
    \label{fig:both-high-shap-beeswarm}
  \end{subfigure}
  \begin{subfigure}
    \includegraphics[width=\linewidth]{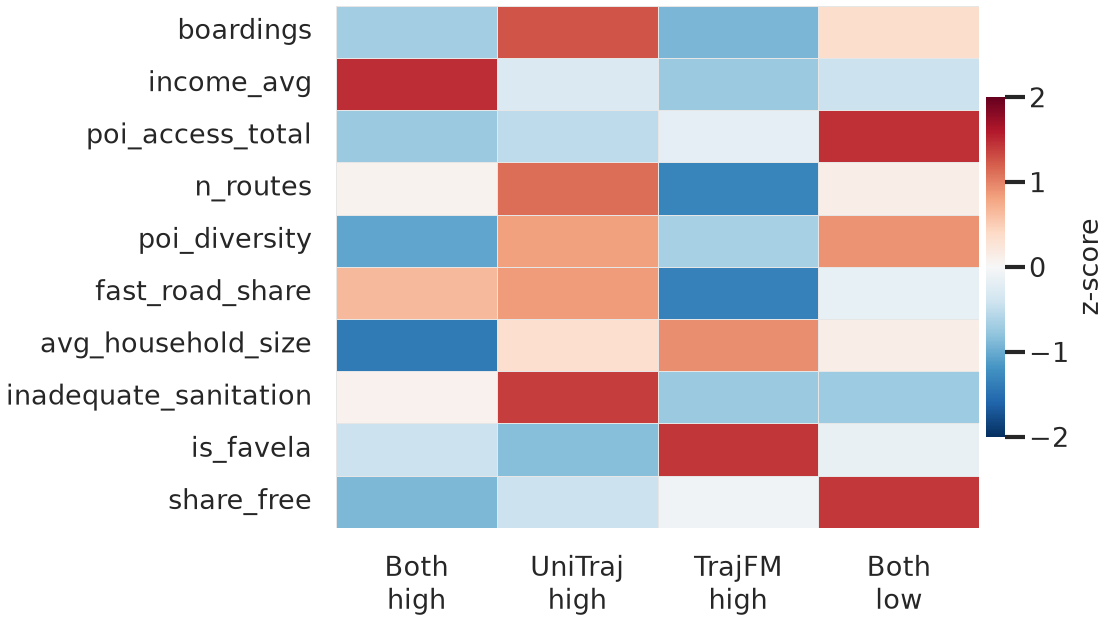}
    \caption{Observed feature profiles across hotspot categories for the top contextual variables (red: above average, blue: below).}
    \label{fig:hotspot-feature-profile}
  \end{subfigure}
  %\label{fig:global_metrics}
\end{figure}

A tuned binary XGBoost classifier distinguishing joint hotspots from all other cells achieved an out-of-fold ROC AUC of $0.87$ and an average precision of $0.46$ against an $11.5\%$ base positive rate. 
%While hard multiclass classification remains modest (balanced accuracy $\approx 0.51$), the framework provides strong hotspot ranking capabilities.
Global SHAP analysis links these combined hotspots to explanatory features. 
Fig.~\ref{fig:both-high-shap-beeswarm} displays the joint hotspot SHAP profile:  hotspots remain strongly associated with passenger pass composition, population density, and educational accessibility. 
The observed associations differ between the two residual types: UniTraj errors are more strongly associated with geometry and accessibility, whereas TrajFM errors show weaker associations, primarily with passenger and timing-related context.
%This suggests that the residual types reflect related but distinct phenomena: UniTraj errors are more strongly associated with geometry and accessibility, whereas TrajFM errors are more closely associated with passenger and timing-related context.

Joint hotspots are distinguished primarily by low population density (Kruskal-Wallis H=27.84, p<0.0001 across the four hotspot categories) rather than income (H=6.07, p=0.11, not significant) — suggesting a stronger association with low-density peripheral urban structure than with income deprivation.
Fig.~\ref{fig:hotspot-feature-profile} illustrates how further top non-speed and non-population variables systematically differ across hotspot classes.

%Crucially, our framework maps these drivers locally to show that the same hotspot class can arise from highly divergent micro-environments. 
%Fig.~\ref{fig:both-high-driver-map} displays the dominant feature assigned to each cell based on its largest absolute SHAP contribution. 
%For instance, where population density or 15-minute accessibility variables (such as education and shopping POI density or overall POI diversity) dominate locally, it isolates areas where dense urban friction disrupts generic movement priors. 
%Conversely, a locally dominant share of free, senior, or cash rides points to specific neighborhoods where targeted ticketing infrastructure, fare-media distribution, or adapted stop designs could best reduce boarding friction. 
%Ultimately, these joint hotspots are not merely foundation model failures: they mark specific windows where bus movement is fundamentally hard to predict from generic mobility priors, identifying exactly where localized operational validation is most warranted.
Fig.~\ref{fig:both-high-driver-map} maps each cell’s dominant SHAP driver: population density, 15-minute accessibility, and fare composition, highlighting areas where generic mobility priors fit poorly and targeted operational validation is most needed.

\section{Conclusion and Future Work}
%Our preliminary experiments demonstrate that transferred trajectory foundation models can provide informative proxy signals of transit irregularity.
%Transferred trajectory foundation models produced spatially structured, interpretable residuals in Niterói’s bus network. 
Transferred trajectory foundation models produced spatially structured, interpretable residuals that may serve as proxy signals of transit irregularity in Niterói’s bus network.

UniTraj and TrajFM outperformed local baselines, while their overlapping high-error cells were consistently associated with population density, accessibility, and passenger composition. 
These residuals should be interpreted as candidate irregularity indicators rather than verified disruptions, but they provide a useful way to prioritize operational investigation. 
Future work will validate hotspots against route-normalized variability and agency records and assess performance under spatially blocked cross-validation.

%Potential avenues for future work include route-normalized irregularity validation, testing spatially blocked cross validation, optimizing spatial aggregation scale, integrating schedules and route layouts, and enhancing foundational models. 
%We also plan to explore explicitly incorporating dynamic meteorological and temporal features. 
%Finally, validating the identified high-error hotspots against known bottlenecks and ground-truth transit agency logs remains a key intended next step.

%\section*{Funding}
%No external funding was received for this work.

%\section*{Acknowledgment}
%The authors thank the NetMob 2026 Data Challenge organizers for preparing and documenting the Niter\'oi public-transport dataset.

%\section*{Disclosures}
%The authors declare no conflicts of interest.

%\section*{Data availability}
%Data analyzed in this paper are available through the NetMob 2026 Data Challenge under its access conditions~\cite{netmob26}.
%Derived aggregate outputs and analysis code are available from the authors upon request.

\bibliography{references}

\end{document}